\documentclass[11pt,a4paper]{article}
\usepackage[hyperref]{acl2018}
\usepackage{times}
\usepackage{latexsym}
\usepackage{booktabs}
\usepackage{url}

\usepackage{verbatim}
\usepackage{amsmath}
\usepackage{cite}
\usepackage{multirow}
\usepackage{graphicx}
\usepackage{float}
\usepackage{subcaption}
\usepackage{enumerate}

\newcommand{\ft}{{\sc{FastText}}}
\newcommand{\wtov}{{\sc{Word2Vec}}}

\newcommand{\pft}{{\sc{pft}}}

\newcommand{\xxcomment}[4]{\textcolor{#1}{[$^{\textsc{#2}}_{\textsc{#3}}$ #4]}}
\newcommand{\ben}[1]{\xxcomment{green}{B}{A}{#1}}

\newcommand{\Nor}{\mathcal{N}}
\renewcommand{\footnotesize}{\tiny}

\aclfinalcopy 
\def\aclpaperid{187} 

\title{Probabilistic FastText for Multi-Sense Word Embeddings}

\author{Ben Athiwaratkun\thanks{ \ \ \ Work done partly during internship at Amazon.}  \\
  Cornell University \\
  {\tt pa338@cornell.edu} \\\And
  Andrew Gordon Wilson \\
  Cornell University \\
  {\tt andrew@cornell.edu} \\ \And
   Anima Anandkumar \\
  AWS \& Caltech \\
  {\tt anima@amazon.com} 
  }

\date{}

\begin{document}
\maketitle
\begin{abstract}

We introduce \emph{Probabilistic FastText}, a new model for word embeddings that can capture multiple word senses, sub-word structure, and uncertainty information.  In particular, we represent each word with a Gaussian mixture density, where the mean of a mixture component is given by the sum of n-grams.  This representation allows the model to share statistical strength across sub-word structures (e.g.\ Latin roots), producing accurate representations of rare, misspelt, or even unseen words. 
Moreover, each component of the mixture can capture a different word sense. Probabilistic FastText outperforms both \ft, which has no probabilistic model, and dictionary-level probabilistic embeddings, which do not incorporate subword structures, on several word-similarity benchmarks, including English RareWord and foreign language datasets.
We also achieve state-of-art performance on benchmarks that measure ability to discern different meanings. 
Thus, the proposed model is the first to achieve multi-sense representations while having enriched semantics on rare words.
\end{abstract}

\section{Introduction}

Word embeddings are foundational to natural language processing.  In order to model language, we need word  representations to contain as much semantic information as possible.  Most research has focused on vector word embeddings, such as \wtov \text{ } \citep{word2vec}, where words with similar meanings are mapped to nearby points in a vector space. Following the seminal work of \citet{word2vec}, there have been numerous works looking to learn efficient word embeddings. 

One shortcoming with the above approaches to word embedding that are based on a predefined dictionary (termed as dictionary-based embeddings) is their inability to learn representations of rare words. To overcome this limitation, character-level word embeddings have been proposed. \ft \ \citep{fasttext} is the state-of-the-art character-level approach to embeddings.  In \ft, each word is modeled by a sum of vectors, with each vector representing an n-gram.  The benefit of this approach is that the training process    can then share {\em strength} across words composed of common roots.  For example, with individual representations for ``circum'' and ``navigation'', we can construct an informative representation for ``circumnavigation'', which would otherwise appear too infrequently to learn a dictionary-level embedding.  In addition to effectively modelling rare words, character-level embeddings can also represent slang or misspelled words, such as ``dogz'', and can share strength across different languages that share roots, e.g.\ Romance languages share latent roots.

A different promising  direction involves representing words with probability distributions, instead of point vectors. For example, \citet{word2gauss} represents words with Gaussian distributions, which can capture uncertainty information.  \citet{word2gm} generalizes this approach to multimodal probability distributions, which can naturally represent words with different meanings. For example, the distribution for ``rock'' could have mass near the word ``jazz'' and ``pop'', but also ``stone'' and ``basalt''.  \citet{doe} further developed this approach to learn hierarchical word representations: for example, the word ``music'' can be learned to have a broad distribution, which encapsulates the distributions for ``jazz'' and ``rock''.  

In this paper, we propose \emph{Probabilistic FastText} (\pft), which provides probabilistic character-level representations of words.  The resulting word embeddings are highly expressive, yet straightforward and interpretable, with simple, efficient, and intuitive training procedures.  \pft \ can model rare words, uncertainty information, hierarchical representations, and multiple word senses.  In particular, we represent each word with a Gaussian or a Gaussian mixture density, which we name  {\sc pft-g} and {\sc pft-gm} respectively. 
Each component of the mixture can represent different word senses, and the mean vectors of each component decompose into vectors of n-grams, to capture character-level information.  We also derive an efficient energy-based max-margin training procedure for \pft.

We perform comparison with \ft \ as well as existing density word embeddings {\sc w2g} (Gaussian) and {\sc w2gm} (Gaussian mixture). 
Our models extract high-quality semantics based on multiple word-similarity benchmarks, including the rare word dataset. 
We obtain an average weighted improvement of $3.7 \%$ over  \ft \ \citep{fasttext} and  $3.1\%$ over the dictionary-level density-based models.  
We also observe meaningful nearest neighbors, particularly in the multimodal density case, where each mode captures a distinct meaning. 
Our models are also directly portable to foreign languages without any hyperparameter modification, where we observe strong performance, outperforming \ft \ on many foreign word similarity datasets. 
Our multimodal word representation can also disentangle meanings, and is able to separate different senses in foreign polysemies. In particular, our models attain state-of-the-art performance on SCWS, a benchmark to measure the ability to separate different word meanings, achieving $1.0\%$ improvement over a recent density embedding model {\sc w2gm} \citep{word2gm}. 

To the best of our knowledge, we are the first to develop multi-sense embeddings with high semantic quality for rare words. Our code and embeddings are publicly available. \footnote{\footnotesize \url{https://github.com/benathi/multisense-prob-fasttext}}

\section{Related Work} \label{sec:related}
Early word embeddings which capture semantic information include \citet{nnlm}, \citet{collobert_we}, and \citet{rnnlm2}.
Later, \citet{word2vec} developed the popular \wtov \ method, which proposes a log-linear model and negative sampling approach that efficiently extracts rich semantics from text.  
Another popular approach {\sc GloVe} learns word embeddings by factorizing co-occurrence matrices \citep{glove}.

Recently there has been a surge of interest in making dictionary-based word embeddings more flexible. This flexibility  has valuable applications in many end-tasks such as  language modeling \citep{char_lm}, named entity recognition \citep{char_ner}, and machine translation \citep{char_mt_efficient, char_mt_noseg}, where unseen words are frequent and proper handling of these words can greatly improve the performance. These works focus on modeling subword information in neural networks for tasks such as language modeling.

Besides vector embeddings, there is recent work on multi-prototype embeddings where each word is represented by multiple vectors. 
The learning approach involves using a cluster centroid of context vectors \citep{multipleprototypes}, or adapting the skip-gram model to learn multiple latent representations \citep{multi_word_embs}. \citet{nonparam_multiprototype} furthers adapts skip-gram with a non-parametric approach to learn the embeddings with  an arbitrary number of senses per word. \citet{unified_sense_chen14} incorporates an external dataset {\sc WordNet} to learn sense vectors. We compare these models with our multimodal  embeddings in Section \ref{sec:evaluation}.

\section{Probabilistic FastText} \label{sec:model}

We introduce \emph{Probabilistic FastText}, which combines a probabilistic word representation with the ability to capture subword structure.  We describe the probabilistic subword representation in Section~\ref{sec:word2gm}.  We then describe the similarity measure and the loss function used to train the embeddings in Sections~\ref{sec:sim_measure} and \ref{sec:loss_function}.  We conclude by briefly presenting a simplified version of the energy function for isotropic Gaussian representations (Section~\ref{sec: energysimp}), and the negative sampling scheme we use in training (Section~\ref{sec:word_sampling}).

\begin{figure}[h]
  \centering
    \begin{minipage}{0.14\textwidth}
  \captionsetup{type=figure}
       \centering
       \centering
       \includegraphics[clip, trim={300 260 330 180},width=1.0\linewidth]{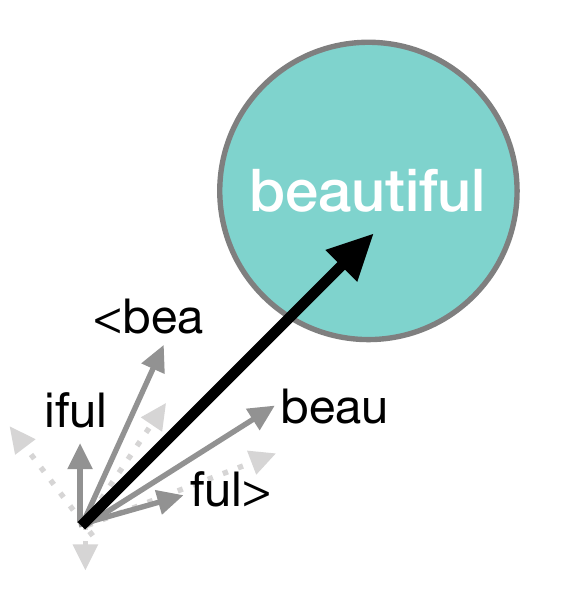}
  \subcaption{ \label{fig:subword_gauss} }
\end{minipage} 
\hspace{0.0\textwidth} 
       \minipage{0.3\textwidth}
       \centering
       \centering
       \includegraphics[clip, trim={180 220 180 180},width=1.0\linewidth]{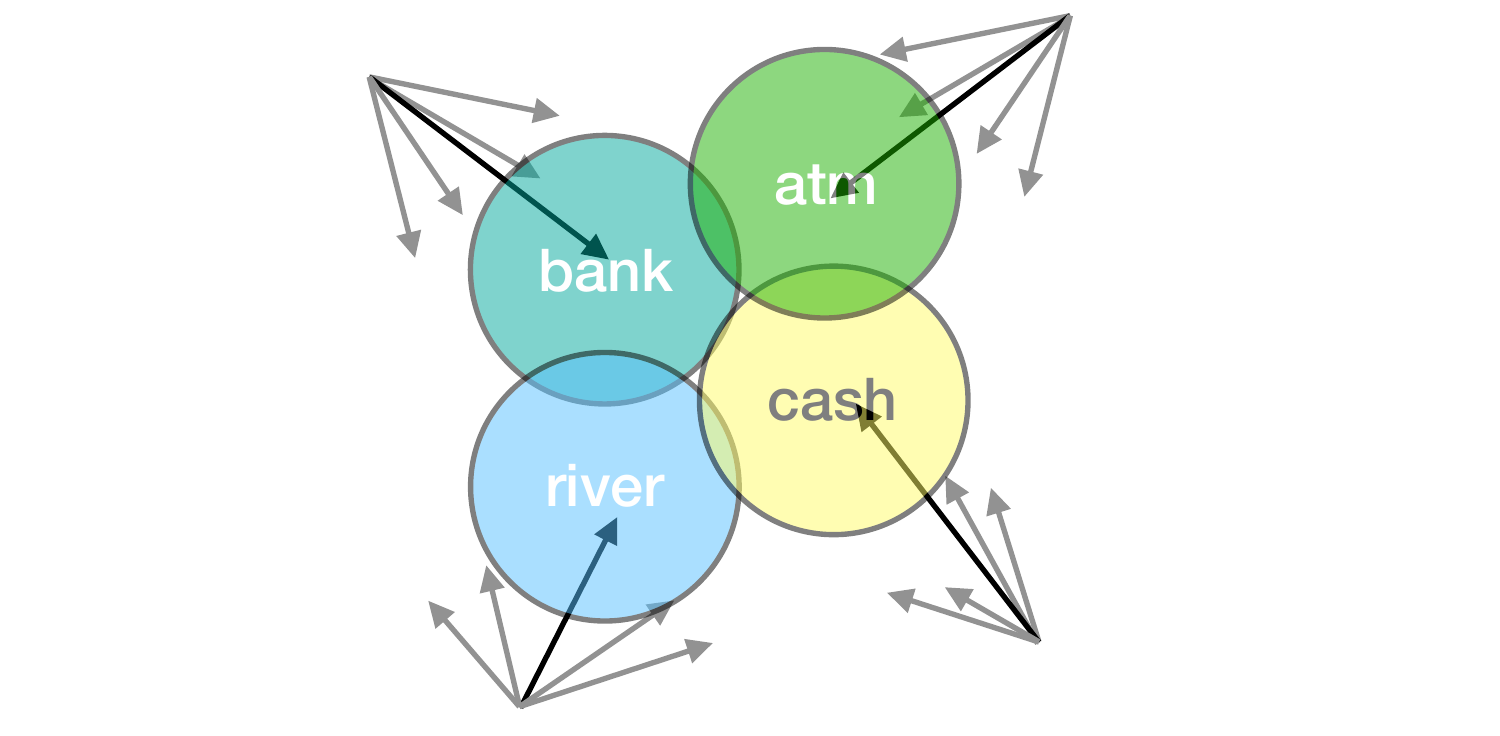}
 \subcaption{\label{fig:c2g}}
\endminipage\hfill
\minipage{0.3\textwidth}
       \centering
       \centering
       \includegraphics[clip, trim={220 250 160 200},width=1.0\linewidth]{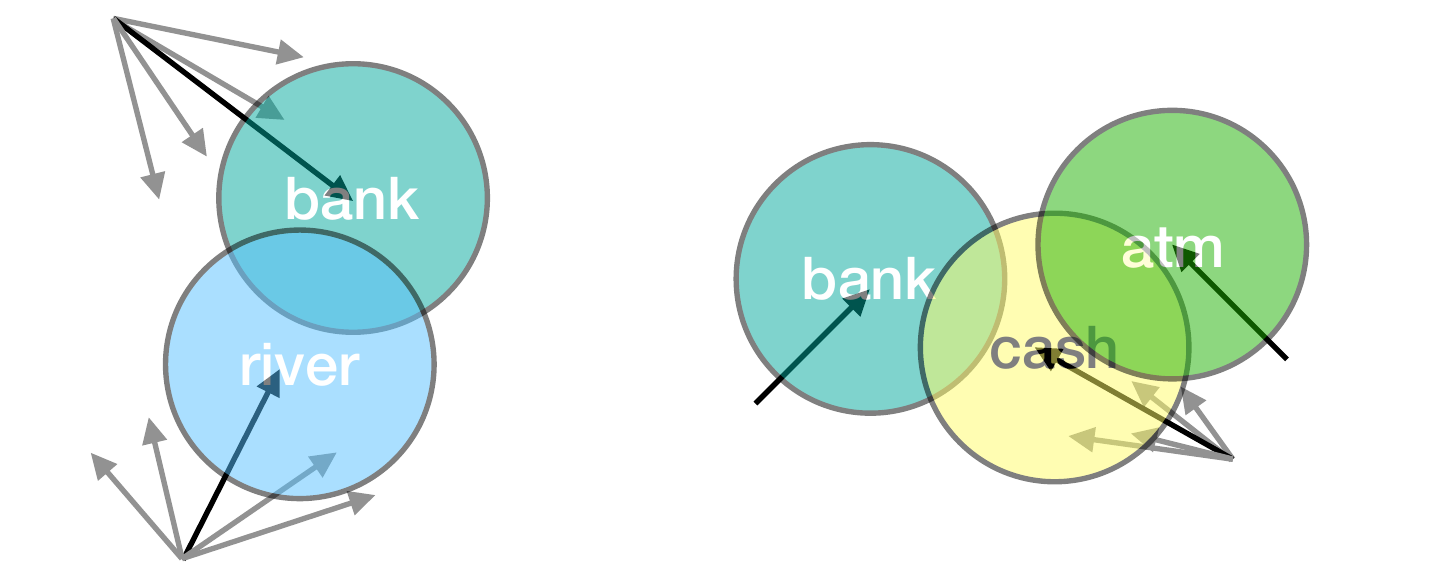}
 \subcaption{\label{fig:c2gm}}
\endminipage\hfill
  \caption{
  (\ref{fig:subword_gauss}) a Gaussian component and its subword structure. The bold arrow represents the final mean vector, estimated from averaging the grey n-gram vectors. 
  (\ref{fig:c2g}) {\sc pft-g } model: Each Gaussian component's mean vector is a subword vector.  
  (\ref{fig:c2gm}) {\sc pft-gm} model: For each Gaussian mixture distribution, one component's mean vector is estimated by a subword structure whereas other components are dictionary-based vectors. 
  }
  \label{fig:c2g_c2gm}
\end{figure}

\subsection{Probabilistic Subword Representation} \label{sec:word2gm}

We represent each word with a Gaussian mixture with $K$ Gaussian components. That is, a  word $w$ is associated with a density function $f(x) = \sum_{i=1}^K p_{w,i} \Nor(x; \vec{\mu}_{w,i} , \Sigma_{w,i} )$ where $\{ \mu_{w,i} \}_{k=1}^K$ are the mean vectors and $\{\Sigma_{w,i}\}$ are the covariance matrices, and $\{p_{w,i}\}_{k=1}^K $ are the component probabilities which sum to $1$. 

The mean vectors of Gaussian components hold much of the semantic information in density embeddings. While these models are successful based on word similarity and entailment benchmarks \citep{word2gauss, word2gm}, the mean vectors are often dictionary-level, which can lead to poor semantic estimates for rare words, or the inability to handle words outside the training corpus. We propose using subword structures to estimate the mean vectors. We outline the formulation below. 

For word $w$,  we estimate the mean vector $\mu_w$ with the average over n-gram vectors and its dictionary-level vector. That is,
\begin{equation} \label{eq:fasttext_subword}
\mu_w = \frac{1}{|NG_w| + 1 } \left(  v_w  + \sum_{g \in NG_w } z_g \right)
\end{equation}
where $z_g$ is a vector associated with an n-gram $g$, $v_w$ is the dictionary representation of word $w$, and $NG_w$ is a set of $n$-grams of word $w$. Examples of 3,4-grams for a word ``beautiful'', including the beginning-of-word character `$\langle$' and end-of-word character `$\rangle$', are: 
\begin{itemize}
\item 3-grams: $\langle$be, bea, eau, aut, uti, tif, ful, ul$\rangle$
\item 4-grams: $\langle$bea, beau .., iful ,ful$\rangle$
\end{itemize}
This structure is similar to that of \ft \ \citep{fasttext}; however, we note that \ft \ uses single-prototype deterministic embeddings as well as a training approach that maximizes the negative log-likelihood, whereas we use a multi-prototype probabilistic embedding and for training we maximize the similarity between the words' probability densities, as described in Sections \ref{sec:sim_measure} and \ref{sec:loss_function}

Figure \ref{fig:subword_gauss} depicts the subword structure for the mean vector. Figure \ref{fig:c2g} and \ref{fig:c2gm} depict our models, Gaussian probabilistic \ft \ ({\sc pft-g}) and Gaussian mixture probabilistic \ft \ ({\sc pft-gm}). In the Gaussian case, we represent each mean vector with a subword estimation. For the Gaussian mixture case, we represent one Gaussian component's mean vector with the subword structure whereas other components' mean vectors are dictionary-based. 
This model choice to use dictionary-based mean vectors for other components is to reduce to constraint imposed by the subword structure and promote independence for meaning discovery.

\subsection{Similarity Measure between Words} \label{sec:sim_measure}

\begin{figure}[h]
  \centering
       \includegraphics[clip, trim={170 320 160 360},width=1.0\linewidth]{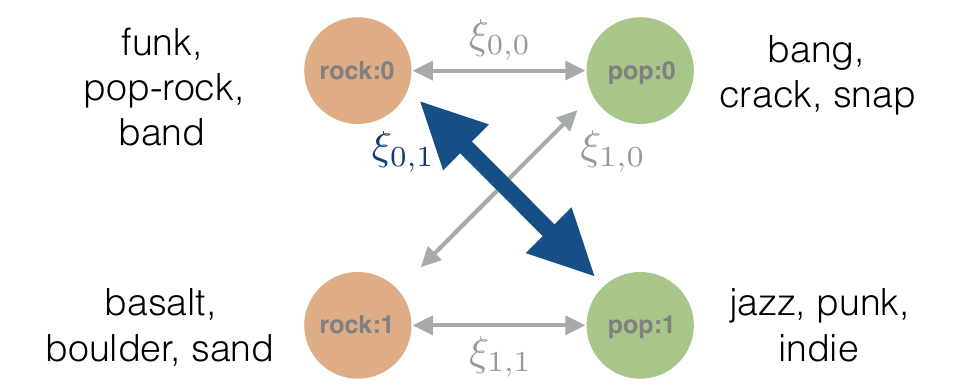}
  \caption{The interactions among Gaussian components of word {\tt rock} and word {\tt pop}. The partial energy is the highest for the pair {\tt rock:0} (the zeroth component of rock) and {\tt pop:1} (the first component of pop), reflecting the similarity in meanings.
}
  \label{fig:partial_energy}
\end{figure}

Traditionally, if words are represented by vectors, a common similarity metric is a dot product. In the case where words are represented by distribution functions, we use the generalized dot product in Hilbert space $\langle \cdot, \cdot \rangle_{L_2}$, which is called the expected likelihood kernel \citep{prob_product_kernel}. We define the energy $E(f,g)$ between two words $f$ and $g$ to be 
$ E (f,g) = \log \langle f, g \rangle_{L_2} = \log  \int f(x) g(x) \ dx$. 
With Gaussian mixtures $f(x) = \sum_{i=1}^K p_i \Nor(x; \vec{\mu}_{f,i} , \Sigma_{f,i} ) $ and $g(x)  =  \sum_{i=1}^K q_i \Nor(x; \vec{\mu}_{g,i} , \Sigma_{g,i} )$, $\sum_{i =1}^K p_i = 1 $, and $\sum_{i =1}^K q_i = 1$, the energy  has a closed form: 
\begin{equation} \label{eq:loge}
  E(f,g) = \log \sum_{j=1}^K \sum_{i=1}^K p_i q_j e^{\xi_{i,j}}
\end{equation}
where $\xi_{j,j}$ is the partial energy which corresponds to the similarity between component $i$ of the first word $f$ and component $j$ of the second word $g$.\footnote{The orderings of indices of the components for each word are arbitrary.} 
\begin{align}
\nonumber
\xi_{i,j} &\equiv \log  \Nor(0; \vec{\mu}_{f,i} - \vec{\mu}_{g,j}, \Sigma_{f,i} + \Sigma_{g,j} ) \\ \nonumber
&= - \frac{1}{2} \log \det( \Sigma_{f,i} + \Sigma_{g,j} ) - \frac{D}{2} \log (2 \pi)  \\ 
  - \frac{1}{2} & (\vec{\mu}_{f,i} - \vec{\mu}_{g,j} )^\top (\Sigma_{f,i} + \Sigma_{g,j} )^{-1} (\vec{\mu}_{f,i} - \vec{\mu}_{g,j} )  \label{eq:partial_energy}
 \end{align}
Figure \ref{fig:partial_energy} demonstrates the partial energies among the Gaussian components of two words. 

\subsection{Loss Function} \label{sec:loss_function}
The model parameters that we seek to learn are $v_w$  for each word $w$ and $z_g$ for each n-gram $g$. We train the model by pushing the energy of a true context pair $w$ and $c$ to be higher than the negative context pair $w$ and $n$ by a margin $m$. We use Adagrad \citep{adagrad} to minimize the following loss to achieve this outcome:
\begin{align} \label{eq:rank_loss}
L(f,g) = \max\left[ 0, m - E(f,g) + E(f,n) \right].
\end{align}
We  describe how to sample words as well as its positive and negative contexts in Section \ref{sec:word_sampling}.

This loss function together with the Gaussian mixture model with $K>1$ has the ability to extract multiple senses of words. That is, for a word with multiple meanings, we can observe each mode to represent a distinct meaning. For instance, one density mode of ``star'' is close to the densities of ``celebrity'' and ``hollywood'' whereas another mode of ``star'' is near the densities of ``constellation'' and ``galaxy''.

\subsection{Energy Simplification} \label{sec: energysimp} 
In theory, it can be beneficial to have covariance matrices as learnable parameters. In practice, \citet{word2gm} observe that spherical covariances often perform on par with diagonal covariances with much less computational resources. Using spherical covariances for each component, we can further simplify the energy function as follows:

\begin{equation} \label{eq:partial_energy_fixvar}
\xi_{i,j} = - \frac{\alpha}{2} \cdot || \mu_{f,i} - \mu_{g,j} ||^2 \,,
\end{equation}
where the hyperparameter $\alpha$ is the scale of the inverse covariance term in Equation \ref{eq:partial_energy}.
We note that Equation \ref{eq:partial_energy_fixvar} is equivalent to Equation \ref{eq:partial_energy}  up to an additive constant given that the covariance matrices are spherical and the same for all components.

\subsection{Word Sampling}   \label{sec:word_sampling}
To generate a context word $c$ of a given word $w$, we pick a nearby word within a context window of a fixed length $\ell$.
We also use a word sampling technique similar to \citet{word2vec2}. 
This subsampling procedure selects words for training with lower probabilities if they appear frequently. 
This technique has an effect of reducing the importance of words such as `the', `a', `to' which can be predominant in a text corpus but are not as meaningful as other less frequent words such as `city', `capital', `animal', etc. 
In particular, word $w$ has probability $P(w) = 1 - \sqrt{t/f(w)}$ where $f(w)$ is the frequency of word $w$ in the corpus and $t$ is the frequency threshold. 

A negative context word is selected using a distribution $P_n(w) \propto U(w)^{3/4}$ where $U(w)$ is a unigram probability of word $w$. The exponent $3/4$ also diminishes the importance of frequent words and shifts the training focus to other less frequent words.

\section{Experiments} \label{sec:evaluation}
We have proposed a probabilistic \ft \ model which combines the flexibility of subword structure with the density embedding approach.
In this section, we show that our probabilistic representation with subword mean vectors with the simplified energy function outperforms many word similarity baselines and provides disentangled meanings for polysemies. 

First, we describe the training details in Section \ref{sec:train_details}. We provide qualitative evaluation in Section \ref{sec:nn}, showing meaningful nearest neighbors for the Gaussian embeddings, as well as the ability to capture multiple meanings by Gaussian mixtures. Our quantitative evaluation in Section \ref{sec:wordsim} demonstrates strong performance against the baseline models \ft \ \citep{fasttext} and the dictionary-level Gaussian ({\sc w2g}) \citep{word2gauss} and Gaussian mixture embeddings \citep{word2gm} ({\sc w2gm}). We train our models on foreign language corpuses and show competitive results on foreign word similarity benchmarks in Section \ref{sec:foreign_eval}. Finally, we explain the importance of the n-gram structures for semantic sharing in Section \ref{sec:subword_decomp}. 

\subsection{Training Details} \label{sec:train_details}

We train our models on both English and foreign language datasets. For English, we use the concatenation of {\sc ukWac} and {\sc WackyPedia} \citep{wacky} which consists of $3.376$ billion words. We filter out word types that occur fewer than $5$ times which results in a vocabulary size of 2,677,466.

For foreign languages, we demonstrate the training of our model on French, German, and Italian text corpuses. We note that our model should be applicable for  other languages as well. We use {\sc FrWac} (French), {\sc DeWac} (German), {\sc ItWac} (Italian) datasets \citep{wacky} for text corpuses, consisting of $1.634$, $1.716$ and $1.955$ billion words respectively. We use the same threshold, filtering out words that occur less than $5$ times in each corpus. We have dictionary sizes of $1.3$, $2.7$, and $1.4$ million words for {\sc FrWac}, {\sc DeWac}, and {\sc ItWac}.

We adjust the hyperparameters on the English corpus and use them for foreign languages. 
 Note that the adjustable parameters for our models are the loss margin $m$ in Equation \ref{eq:rank_loss} and the scale $\alpha$ in Equation \ref{eq:partial_energy_fixvar}.  We search for the optimal hyperparameters in a grid $m \in 
\{ 0.01, 0.1, 1, 10, 100 \}$ and $\alpha \in \{\frac{1}{5 \times 10^{-3}} , \frac{1}{10^{-3}}, \frac{1}{2 \times 10^{-4}}, \frac{1}{1 \times 10^{-4}}  \}$ on our English corpus. 
The hyperpameter $\alpha$ affects the scale of the loss function; therefore, we adjust the learning rate appropriately for each $\alpha$. In particular, the learning rates used are $\gamma = \{10^{-4}, 10^{-5}, 10^{-6} \}$ for the respective $\alpha$ values.

Other fixed hyperparameters include the number of Gaussian components $K=2$, the context window length $\ell = 10$ and the subsampling threshold $t = 10^{-5}$. Similar to the setup in \ft, we use n-grams where $n=3,4,5,6$ to estimate the mean vectors.

\subsection{Qualitative Evaluation - Nearest neighbors} \label{sec:nn}
We show that our embeddings learn the word semantics well by demonstrating meaningful nearest neighbors. Table \ref{table:nn}  shows examples of polysemous words such as {\tt rock}, {\tt star}, and {\tt cell}. 

\begin{table*}[h]
\begin{center}
\begin{small}
\begin{tabular}{ccc}
\toprule
Word & Co. & Nearest Neighbors \\
\midrule
rock	& 0 & rock:0, rocks:0, rocky:0, mudrock:0, rockscape:0, boulders:0 , coutcrops:0, \\
rock 	& 1 & rock:1, punk:0, punk-rock:0, indie:0, pop-rock:0, pop-punk:0, indie-rock:0, band:1 \\ 
bank & 0 & bank:0, banks:0, banker:0, bankers:0, bankcard:0, Citibank:0, debits:0\\
bank & 1 & bank:1, banks:1, river:0, riverbank:0, embanking:0, banks:0, confluence:1 \\
star & 0 &  stars:0, stellar:0, nebula:0, starspot:0, stars.:0, stellas:0, constellation:1\\
star & 1 & star:1, stars:1, star-star:0, 5-stars:0, movie-star:0, mega-star:0, super-star:0\\
cell & 0 &  cell:0, cellular:0, acellular:0, lymphocytes:0, T-cells:0, cytes:0, leukocytes:0\\
cell & 1 & cell:1, cells:1, cellular:0, cellular-phone:0, cellphone:0, transcellular:0\\
left & 0 & left:0, right:1, left-hand:0, right-left:0, left-right-left:0,  right-hand:0,   leftwards:0\\
left & 1 & left:1, leaving:0, leavings:0,  remained:0, leave:1, enmained:0, leaving-age:0, sadly-departed:0 \\
\bottomrule
\end{tabular}

\smallskip
\begin{tabular}{ccc}
\toprule
Word  & Nearest Neighbors \\
\midrule
rock &  rock, rock-y, rockn, rock-, rock-funk, rock/, lava-rock, nu-rock, rock-pop, rock/ice, coral-rock \\
bank & bank-, bank/, bank-account, bank., banky, bank-to-bank, banking, Bank, bank/cash, banks.** \\
star &  movie-stars, star-planet, G-star, star-dust, big-star, starsailor, 31-star, star-lit, Star, starsign, pop-stars\\
cell &  cellular, tumour-cell, in-cell, cell/tumour, 11-cell, T-cell, sperm-cell, 2-cells, Cell-to-cell\\
left & left, left/joined, leaving, left,right, right, left)and, leftsided, lefted, leftside\\
\bottomrule
\end{tabular}
\end{small}
\end{center}
\caption{Nearest neighbors of {\sc pft-gm} (top) and {\sc pft-g} (bottom). The notation {\tt w:i} denotes  the $i^{th}$ mixture component of the word {\tt w}.
}
\label{table:nn}
\end{table*}

Table \ref{table:nn} shows the nearest neighbors of polysemous words. We note that  subword embeddings prefer words with overlapping characters as nearest neighbors. For instance, ``rock-y'', ``rockn'', and ``rock'' are both close to the word ``rock''. For the purpose of demonstration, we only show words with meaningful variations and omit words with small character-based variations previously mentioned. However, all words shown are in the top-100 nearest words.

 We observe the separation in meanings for the multi-component case; for instance, one component of the word ``bank'' corresponds to a financial bank whereas the other component corresponds to a river bank. The single-component case also has interesting behavior. We observe that the subword embeddings of polysemous words can represent both meanings. For instance, both ``lava-rock'' and ``rock-pop'' are among the closest words to ``rock''.

\subsection{Word Similarity Evaluation} \label{sec:wordsim}

\begin{table*}[ht!]
\centering
\begin{sc}
\begin{tabular}{c | c c c c | c c c c c c c }
\hline
 \multirow{2}{*}{ {\sc } } 
 	d	& \multicolumn{4}{c}{50} & \multicolumn{5}{| c}{300}  \\  \hline
				& w2g			& 	w2gm		&	pft-g 		& pft-gm		&	\ft 			& w2g		& 	w2gm		&	pft-g 		& pft-gm \\
\midrule	
SL-999 		&	29.35 	&      29.31 		&  27.34  	& \textbf{34.13} 	&  38.03	&	38.84   	&    \textbf{39.62}      &   35.85 				&	39.60
\\WS-353 	& 	71.53 	&      \textbf{73.47} 		&  67.17  	& 71.10 	&	73.88	& 78.25 	&    \textbf{79.38}     	&  73.75   				& 76.11
\\MEN-3k 	&  72.58 	&      73.55 		&  70.61	& \textbf{73.90} 	&	76.37	& 78.40 	&    78.76 		&   77.78  				& \textbf{79.65}
\\MC-30 	&  76.48 	&      79.08		&  73.54	& \textbf{79.75}  	&	81.20	& 82.42  	&    \textbf{84.58}      & 	81.90				& 80.93 
\\RG-65		&  73.30 	&    	74.51 		&  70.43	& \textbf{78.19}  	&	79.98	& 80.34		&    \textbf{80.95} 		& 	77.57				& 79.81
\\YP-130 	&  41.96 	&      \textbf{45.07} 		&  37.10	& 40.91 	&	53.33	& 46.40		&    47.12  	&  48.52				& \textbf{54.93}
\\MT-287	&  64.79 	&      66.60 		&  63.96	& \textbf{67.65}  	&	67.93	& 67.74     &    \textbf{69.65}    	&  66.41	& 69.44
\\MT-771 	&  60.86  	&   	60.82 		&  60.40	& \textbf{63.86}  	&	66.89	& 70.10  	&    \textbf{70.36}   	& 	67.18				& 69.68
\\RW-2k 	&  28.78 	&     	28.62 		&   44.05  	& \textbf{42.78}  	&	48.09	& 35.49  	&    42.73 		&  \textbf{50.37} 				& 49.36 
\\ \hline
avg.			&	42.32	&	42.76			&	44.35	& \textbf{46.47}		& 	49.28	&	47.71	&	49.54		&	49.86				& 	\textbf{51.10}
\\
\end{tabular}
\end{sc}
\caption{Spearman's Correlation $\rho \times 100$ on Word Similarity Datasets. 
}
\label{table:wordsim_eng}
\label{table:ws_results_300}
\label{table:ws_results_50}
\label{table:results_300d}
\end{table*}

We  evaluate our embeddings on several standard word similarity datasets, namely, SL-999 \citep{evaldata_simlex}, WS-353 \citep{evaldata_wordsim}, MEN-3k \citep{evaldata_men}, MC-30 \citep{evaldata_mc}, RG-65 \citep{evaldata_rg}, YP-130 \citep{evaldata_yp}, MTurk(-287,-771) \citep{evaldata_mturk287, evaldata_mturk771}, and RW-2k \citep{rarewords}. 
Each dataset contains a list of word pairs with a human score of how related or similar the two words are. We use the notation {\sc dataset-num} to denote the number of word pairs {\sc num} in each evaluation set.  We note that the dataset RW focuses more on infrequent words and SimLex-999 focuses on the similarity of words rather than relatedness. We also compare {\sc pft-gm} with other multi-prototype embeddings in the literature using SCWS \citep{multipleprototypes}, a word similarity dataset that is aimed to measure the ability of embeddings to discern multiple meanings.

We calculate the Spearman correlation \citep{spearman04} between the labels and our scores generated by the embeddings. The Spearman correlation is a rank-based correlation measure that assesses how well the scores describe the true labels. The scores we use are cosine-similarity scores between the mean vectors. In the case of Gaussian mixtures, we use the pairwise maximum score:
\begin{equation} \label{eq:maxsim}
s(f,g) = \max_{i \in 1, \hdots, K} \max_{j \in 1,\hdots, K}  \frac{ \mu_{f,i} \cdot \mu_{g,j} }{ ||\mu_{f,i}|| \cdot ||\mu_{g,j}||  }.
\end{equation}
The pair $(i,j)$ that achieves the maximum cosine similarity corresponds to the Gaussian component pair that is the closest in meanings. Therefore, this similarity score yields the most  related senses of a given word pair. This score reduces to a cosine similarity in the Gaussian case $(K=1)$.

\subsubsection{Comparison Against Dictionary-Level Density Embeddings and \ft }

We compare our models against the dictionary-level Gaussian and Gaussian mixture embeddings in Table \ref{table:wordsim_eng}, with $50$-dimensional and $300$-dimensional mean vectors. The 50-dimensional results for {\sc w2g} and {\sc w2gm} are obtained directly from \citet{word2gm}. For comparison, we use the public code\footnote{\url{https://github.com/benathi/word2gm}} to train the 300-dimensional {\sc w2g} and {\sc w2gm} models and the publicly available \ft \ model$\footnote{\url{https://s3-us-west-1.amazonaws.com/fasttext-vectors/wiki.en.zip}}$.

We calculate Spearman's correlations for each of the word similarity datasets. These datasets vary greatly in the number of word pairs; therefore, we mark each dataset with its size for visibility. For a fair and objective comparison, we calculate a weighted average of the correlation scores for each model.

Our {\sc pft-gm} achieves the highest average score among all competing models, outperforming both \ft \ and the dictionary-level embeddings {\sc w2g} and {\sc w2gm}. Our unimodal model {\sc pft-g} also outperforms the dictionary-level counterpart {\sc w2g} and \ft. We note that the model {\sc w2gm} appears quite strong according to Table \ref{table:wordsim_eng}, beating {\sc pft-gm} on many word similarity datasets. 
However, the datasets that {\sc w2gm} performs better than {\sc pft-gm} often have small sizes such as MC-30 or RG-65, where the Spearman's correlations are more subject to noise. 
Overall, {\sc pft-gm} outperforms {\sc w2gm} by $3.1 \%$ and $8.7\%$  in $300$ and $50$  dimensional models.
In addition,  {\sc pft-g} and {\sc pft-gm} also outperform \ft \ by $1.2\%$ and $3.7\%$ respectively.

\subsubsection{Comparison Against Multi-Prototype Models} \label{sec:compare_multi}
In Table \ref{table:multi-prototypes}, we compare $50$ and $300$ dimensional  {\sc pft-gm} models against the multi-prototype embeddings described in Section \ref{sec:related} and the existing multimodal density embeddings {\sc w2gm}. We use the word similarity dataset SCWS \citep{multipleprototypes} which contains words with potentially many meanings, and is a benchmark for distinguishing senses. We use the maximum similarity score (Equation \ref{eq:maxsim}), denoted as {\sc MaxSim}. {\sc AveSim} denotes the average of the similarity scores, rather than the maximum.

We outperform the dictionary-based density embeddings {\sc w2gm} in both $50$ and $300$ dimensions, demonstrating the benefits of subword information. Our model achieves state-of-the-art results, similar to that of \citet{nonparam_multiprototype}.

\begin{table}
\begin{tabular}{l| c | c | c | c}
\toprule
Model 		&	Dim		& 	$\rho \times 100$ \\
\midrule
{\sc Huang AvgSim} 		& 	50		& 	62.8 	\\
{\sc Tian MaxSim}			& 	50		&	63.6	\\
{\sc w2gm MaxSim} 			&	50		&  62.7 \\
{\sc Neelakantan AvgSim}		& 	50		&	64.2 \\
{\sc pft-gm MaxSim} 			& 	50 	&  63.7	\\
\midrule 
{\sc Chen-M AvgSim}		&	200	& 66.2 \\
{\sc w2gm	MaxSim}			&	200	& 65.5 \\
\midrule
{\sc Neelakantan AvgSim} &	300	&	\textbf{67.2}	\\
{\sc w2gm MaxSim}			&	300	&	66.5	\\ 
{\sc pft-gm MaxSim}			&	300	&	\textbf{67.2}	\\
\bottomrule
\end{tabular}
\caption{Spearman's Correlation $\rho \times 100$ on word similarity dataset SCWS. 
} \label{table:multi-prototypes}
\end{table}

\begin{table*}[ht!]
\centering
\begin{tabular}{l | r | c c c c c }
\hline
Lang. & Evaluation 	& \ft 				& w2g	& w2gm	& pft-g 		& pft-gm  \\ 
 \hline \multirow{1}{*}{ {\sc fr} } 
 		& {\sc WS353}	& 		38.2		&	16.73	&	20.09	& 	41.0		&	\textbf{41.3}	\\  %
 \hline \multirow{2}{*}{ {\sc de} }  
 		& {\sc GUR350} &		70		&	65.01	&	69.26	&		77.6		&  \textbf{78.2}  	\\
 		& {\sc GUR65} 	 &		81		&	74.94 	&	76.89	&		81.8		& 	\textbf{85.2}		\\ 
 \hline \multirow{ 2}{*}{ {\sc it} } 
 		& {\sc WS353} 	&		57.1 	&	56.02	&	61.09	&		60.2	 	& \textbf{62.5}		\\ 
		& {\sc SL-999} 	&		29.3	&	29.44	&	\textbf{34.91}	&		29.3		& 33.7 \\ %
\end{tabular}
\caption{Word similarity evaluation on foreign languages.}
\label{table:foreign_all}
\end{table*}

\subsection{Evaluation on Foreign Language Embeddings} \label{sec:foreign_eval}
We evaluate the foreign-language embeddings on word similarity datasets in respective languages. We use Italian {\sc WordSim353} and Italian {\sc SimLex-999} \citep{it_de_rus_ws_sl} for Italian models, {\sc GUR350} and {\sc GUR65} \citep{gur_datasets} for German models, and French {\sc WordSim353} \citep{evaldata_wordsim} for French models.  For datasets GUR350 and GUR65, we use the results reported in the \ft \ publication \citep{fasttext}. For other datasets, we train \ft \ models for comparison using the public code\footnote{\url{https://github.com/facebookresearch/fastText.git}} on our text corpuses. We also train dictionary-level models {\sc w2g}, and {\sc w2gm} for comparison.

Table \ref{table:foreign_all} shows the Spearman's correlation results of our models. 
We outperform \ft \ on many  word similarity benchmarks. Our results are also significantly better than the dictionary-based models, {\sc w2g} and {\sc w2gm}. We hypothesize that {\sc w2g} and {\sc w2gm} can perform better than the current reported results given proper pre-processing of words due to special characters such as accents.

We investigate the nearest neighbors of polysemies in foreign languages and also observe clear sense separation. 
For example, \emph{piano} in Italian can mean ``floor'' or ``slow''. These two meanings are reflected in the nearest neighbors where one component is close to \emph{piano-piano}, \emph{pianod} which mean ``slowly'' whereas the other component is close to \emph{piani} (floors), \emph{istrutturazione} (renovation) or \emph{infrastruttre} (infrastructure). Table \ref{table:foreign_nn} shows additional results, demonstrating that the disentangled semantics can be observed in multiple languages.

\begin{table*}[h]
\begin{center}
\begin{small}
\begin{tabular}{l c l l}
\toprule
Word   & Meaning & Nearest Neighbors \\
\midrule
({\sc it}) \emph{secondo} 		& 2nd 				& Secondo (2nd), terzo (3rd) , quinto (5th), primo (first), quarto (4th), ultimo (last)  \\
({\sc it}) \emph{secondo} 	 	& according to 	& conformit (compliance), attenendosi (following), cui (which), conformemente (accordance with) \\ 
({\sc it}) \emph{porta} 		 	& lead, bring		& portano (lead), conduce (leads), portano, porter, portando (bring), costringe (forces) \\ 
({\sc it}) \emph{porta} 		 	& door  				&	porte (doors), finestrella (window), finestra (window), portone (doorway), serratura (door lock) \\ 
({\sc fr}) \emph{voile} 			& veil 				&	voiles (veil), voiler (veil), voilent (veil), voilement, foulard (scarf), voils (veils), voilant (veiling) \\ 
({\sc fr}) \emph{voile}   		 	& sail 				&	catamaran (catamaran), driveur (driver),  nautiques (water), Voile (sail), driveurs (drivers) \\ 
({\sc fr}) \emph{temps} 			& weather 		&	brouillard (fog), orageuses (stormy), nuageux (cloudy) \\
({\sc fr}) \emph{temps}		 	& time 				& 	mi-temps (half-time), partiel (partial), Temps (time), annualis (annualized), horaires (schedule) \\ 
({\sc fr}) \emph{voler} 		 	& steal 				& envoler (fly),  voleuse (thief), cambrioler (burgle), voleur (thief), violer (violate), picoler (tipple) \\
({\sc fr}) \emph{voler} 		 	& fly 					&	airs (air), vol (flight), volent (fly), envoler (flying), atterrir (land) \\
\bottomrule
\end{tabular}
\caption{Nearest neighbors of polysemies based on our foreign language {\sc pft-gm} models.
}
\label{table:foreign_nn}
\end{small}
\end{center}
\end{table*}

\subsection{Qualitative Evaluation - Subword Decomposition} \label{sec:subword_decomp}

One of the motivations for using subword information is the ability to handle out-of-vocabulary words. Another benefit is the ability to help improve the semantics of rare words via subword sharing. Due to an observation that text corpuses follow Zipf's power law \citep{zipf}, words at the tail of the occurrence distribution appears much less frequently. Training these words to have a good semantic representation is challenging if done at the word level alone. However, an n-gram such as `abnorm' is trained during both occurrences of ``abnormal'' and ``abnormality'' in the corpus, hence further augments both words's semantics. 

\begin{figure}[h!]
\begin{center}
\centerline{\includegraphics[width=0.7\columnwidth,trim={0 20 40 40},clip]{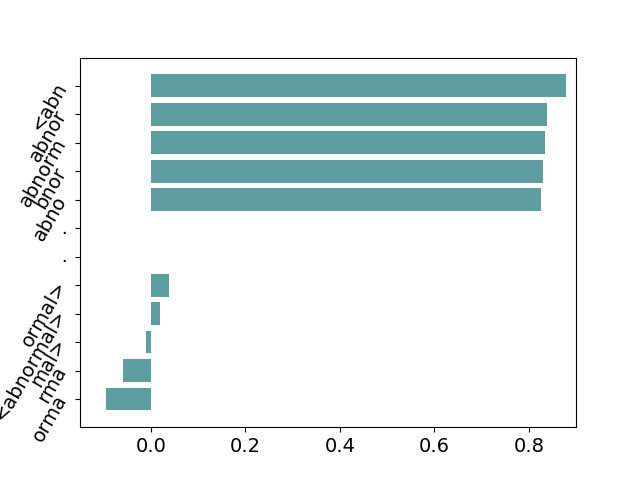}}
\centerline{\includegraphics[width=0.7\columnwidth,trim={0 20 40 40},clip]{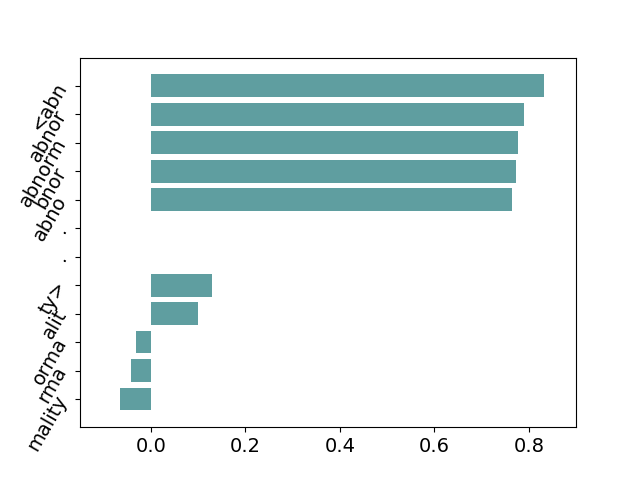}}
\caption{
Contribution of each n-gram vector to the final representation for word ``abnormal'' (top) and  ``abnormality'' (bottom). The x-axis is the cosine similarity between each n-gram vector $z_g^{(w)}$ and the final vector $\mu_w$. 
}
\label{fig:ngram_decomp}
\end{center}
\end{figure}

Figure \ref{fig:ngram_decomp} shows the contribution of  n-grams to the final representation. We filter out to show only the n-grams with the top-5  and bottom-5 similarity scores. We observe that the final representations of both words align with n-grams ``abno'',  ``bnor'', ``abnorm'', ``anbnor'', ``$\textless$abn''. In fact, both ``abnormal'' and ``abnormality'' share the same top-5  n-grams. 
Due to the fact that many rare words such as ``autobiographer'', ``circumnavigations'',  or ``hypersensitivity'' are composed from many common sub-words, the n-gram structure can help improve the representation quality. 

\section{Numbers of Components} \label{sec:discussion}
It is possible to train our approach with $K>2$ mixture components; however, \citet{word2gm} observe that dictionary-level Gaussian mixtures with $K=3$ do not overall improve word similarity results, even though these mixtures can discover $3$ distinct senses for certain words.  Indeed, while $K>2$ in principle allows for greater flexibility than $K=2$, most words can be very flexibly modelled with a mixture of two Gaussians, leading to $K=2$ representing a good balance between flexibility and Occam's razor.

Even for words with single meanings, our \pft\text{ }model with $K=2$ often learns richer representations than a $K=1$ model.  For example, the two mixture components can learn to cluster together to form a more heavy tailed unimodal distribution which captures a word with one dominant meaning but with close relationships to a wide range of other words. 

In addition, we observe that our model with $K$ components can capture more than $K$ meanings. For instance, in $K=1$ model, the word pairs (``cell'', ``jail'') and (``cell'', ``biology'') and (``cell'', ``phone'') will all have positive similarity scores based on $K=1$ model.  In general, if a word has multiple meanings, these meanings are usually compressed into the linear substructure of the embeddings \citep{subword_structure_word_senses}. However, the pairs of non-dominant words often have lower similarity scores, which might not accurately reflect their true similarities.

\section{Conclusion and Future Work}
We have proposed models for probabilistic word representations equipped with flexible sub-word structures, suitable  for rare and out-of-vocabulary words. 
The proposed probabilistic formulation incorporates uncertainty information and naturally allows one to uncover multiple meanings with multimodal density representations.
Our models offer better semantic quality, outperforming competing models on word similarity benchmarks. 
Moreover, our multimodal density models can provide interpretable and disentangled representations, and are the first multi-prototype embeddings that can handle rare words. 

Future work includes an investigation into the trade-off between learning full covariance matrices for each word distribution, computational complexity, and performance. 
This direction can potentially have a great impact on tasks where the variance information is crucial, such as for hierarchical modeling with probability distributions \citep{doe}. 

Other future work involves co-training \pft \ on many languages. 
Currently, existing work on multi-lingual embeddings align the word semantics on pre-trained vectors \citep{multilingual_emb}, which can be suboptimal due to polysemies. 
We envision that the multi-prototype nature can help disambiguate words with multiple meanings and facilitate semantic alignment.

\bibliography{subword_multisense}

\begin{thebibliography}{}
\expandafter\ifx\csname natexlab\endcsname\relax\def\natexlab#1{#1}\fi

\bibitem[{Arora et~al.(2016)Arora, Li, Liang, Ma, and
  Risteski}]{subword_structure_word_senses}
Sanjeev Arora, Yuanzhi Li, Yingyu Liang, Tengyu Ma, and Andrej Risteski. 2016.
\newblock \href{http://arxiv.org/abs/1601.03764}{Linear algebraic structure of
  word senses, with applications to polysemy}.
\newblock {\em CoRR\/} abs/1601.03764.
\newblock
  \href{http://arxiv.org/abs/1601.03764}{http://arxiv.org/abs/1601.03764}.

\bibitem[{Athiwaratkun and Wilson(2017)}]{word2gm}
Ben Athiwaratkun and Andrew~Gordon Wilson. 2017.
\newblock \href{https://arxiv.org/abs/1704.08424}{Multimodal word
  distributions}.
\newblock In {\em ACL\/}.
\newblock
  \href{https://arxiv.org/abs/1704.08424}{https://arxiv.org/abs/1704.08424}.

\bibitem[{Athiwaratkun and Wilson(2018)}]{doe}
Ben Athiwaratkun and Andrew~Gordon Wilson. 2018.
\newblock On modeling hierarchical data via probabilistic order embeddings.
\newblock {\em ICLR\/} .

\bibitem[{Baroni et~al.(2009)Baroni, Bernardini, Ferraresi, and
  Zanchetta}]{wacky}
Marco Baroni, Silvia Bernardini, Adriano Ferraresi, and Eros Zanchetta. 2009.
\newblock \href{https://doi.org/10.1007/s10579-009-9081-4}{The wacky wide web:
  a collection of very large linguistically processed web-crawled corpora}.
\newblock {\em Language Resources and Evaluation\/} 43(3):209--226.
\newblock
  \href{https://doi.org/10.1007/s10579-009-9081-4}{https://doi.org/10.1007/s10579-009-9081-4}.

\bibitem[{Bengio et~al.(2003)Bengio, Ducharme, Vincent, and Janvin}]{nnlm}
Yoshua Bengio, R{\'{e}}jean Ducharme, Pascal Vincent, and Christian Janvin.
  2003.
\newblock \href{http://www.jmlr.org/papers/v3/bengio03a.html}{A neural
  probabilistic language model}.
\newblock {\em Journal of Machine Learning Research\/} 3:1137--1155.
\newblock
  \href{http://www.jmlr.org/papers/v3/bengio03a.html}{http://www.jmlr.org/papers/v3/bengio03a.html}.

\bibitem[{Bojanowski et~al.(2016)Bojanowski, Grave, Joulin, and
  Mikolov}]{fasttext}
Piotr Bojanowski, Edouard Grave, Armand Joulin, and Tomas Mikolov. 2016.
\newblock \href{http://arxiv.org/abs/1607.04606}{Enriching word vectors with
  subword information}.
\newblock {\em CoRR\/} abs/1607.04606.
\newblock
  \href{http://arxiv.org/abs/1607.04606}{http://arxiv.org/abs/1607.04606}.

\bibitem[{Bruni et~al.(2014)Bruni, Tran, and Baroni}]{evaldata_men}
Elia Bruni, Nam~Khanh Tran, and Marco Baroni. 2014.
\newblock \href{http://dl.acm.org/citation.cfm?id=2655713.2655714}{Multimodal
  distributional semantics}.
\newblock {\em J. Artif. Int. Res.\/} 49(1):1--47.
\newblock
  \href{http://dl.acm.org/citation.cfm?id=2655713.2655714}{http://dl.acm.org/citation.cfm?id=2655713.2655714}.

\bibitem[{Chen et~al.(2014)Chen, Liu, and Sun}]{unified_sense_chen14}
Xinxiong Chen, Zhiyuan Liu, and Maosong Sun. 2014.
\newblock \href{http://aclweb.org/anthology/D/D14/D14-1110.pdf}{A unified model
  for word sense representation and disambiguation}.
\newblock In {\em Proceedings of the 2014 Conference on Empirical Methods in
  Natural Language Processing, {EMNLP} 2014, October 25-29, 2014, Doha, Qatar,
  {A} meeting of SIGDAT, a Special Interest Group of the {ACL}\/}. pages
  1025--1035.
\newblock
  \href{http://aclweb.org/anthology/D/D14/D14-1110.pdf}{http://aclweb.org/anthology/D/D14/D14-1110.pdf}.

\bibitem[{Collobert and Weston(2008)}]{collobert_we}
Ronan Collobert and Jason Weston. 2008.
\newblock \href{http://doi.acm.org/10.1145/1390156.1390177}{A unified
  architecture for natural language processing: deep neural networks with
  multitask learning}.
\newblock In {\em Machine Learning, Proceedings of the Twenty-Fifth
  International Conference {(ICML} 2008), Helsinki, Finland, June 5-9, 2008\/}.
  pages 160--167.
\newblock
  \href{http://doi.acm.org/10.1145/1390156.1390177}{http://doi.acm.org/10.1145/1390156.1390177}.

\bibitem[{Duchi et~al.(2011)Duchi, Hazan, and Singer}]{adagrad}
John~C. Duchi, Elad Hazan, and Yoram Singer. 2011.
\newblock \href{http://dl.acm.org/citation.cfm?id=2021068}{Adaptive subgradient
  methods for online learning and stochastic optimization}.
\newblock {\em Journal of Machine Learning Research\/} 12:2121--2159.
\newblock
  \href{http://dl.acm.org/citation.cfm?id=2021068}{http://dl.acm.org/citation.cfm?id=2021068}.

\bibitem[{Finkelstein et~al.(2002)Finkelstein, Gabrilovich, Matias, Rivlin,
  Solan, Wolfman, and Ruppin}]{evaldata_wordsim}
Lev Finkelstein, Evgeniy Gabrilovich, Yossi Matias, Ehud Rivlin, Zach Solan,
  Gadi Wolfman, and Eytan Ruppin. 2002.
\newblock \href{http://doi.acm.org/10.1145/503104.503110}{Placing search in
  context: the concept revisited}.
\newblock {\em {ACM} Trans. Inf. Syst.\/} 20(1):116--131.
\newblock
  \href{http://doi.acm.org/10.1145/503104.503110}{http://doi.acm.org/10.1145/503104.503110}.

\bibitem[{Gurevych(2005)}]{gur_datasets}
Iryna Gurevych. 2005.
\newblock Using the structure of a conceptual network in computing semantic
  relatedness.
\newblock In {\em Natural Language Processing - {IJCNLP} 2005, Second
  International Joint Conference, Jeju Island, Korea, October 11-13, 2005,
  Proceedings\/}. pages 767--778.

\bibitem[{Halawi et~al.(2012)Halawi, Dror, Gabrilovich, and
  Koren}]{evaldata_mturk771}
Guy Halawi, Gideon Dror, Evgeniy Gabrilovich, and Yehuda Koren. 2012.
\newblock \href{http://doi.acm.org/10.1145/2339530.2339751}{Large-scale
  learning of word relatedness with constraints}.
\newblock In {\em The 18th {ACM} {SIGKDD} International Conference on Knowledge
  Discovery and Data Mining, {KDD} '12, Beijing, China, August 12-16, 2012\/}.
  pages 1406--1414.
\newblock
  \href{http://doi.acm.org/10.1145/2339530.2339751}{http://doi.acm.org/10.1145/2339530.2339751}.

\bibitem[{Hill et~al.(2014)Hill, Reichart, and Korhonen}]{evaldata_simlex}
Felix Hill, Roi Reichart, and Anna Korhonen. 2014.
\newblock \href{http://arxiv.org/abs/1408.3456}{Simlex-999: Evaluating semantic
  models with (genuine) similarity estimation}.
\newblock {\em CoRR\/} abs/1408.3456.
\newblock
  \href{http://arxiv.org/abs/1408.3456}{http://arxiv.org/abs/1408.3456}.

\bibitem[{Huang et~al.(2012)Huang, Socher, Manning, and
  Ng}]{multipleprototypes}
Eric~H. Huang, Richard Socher, Christopher~D. Manning, and Andrew~Y. Ng. 2012.
\newblock \href{http://www.aclweb.org/anthology/P12-1092}{Improving word
  representations via global context and multiple word prototypes}.
\newblock In {\em The 50th Annual Meeting of the Association for Computational
  Linguistics, Proceedings of the Conference, July 8-14, 2012, Jeju Island,
  Korea - Volume 1: Long Papers\/}. pages 873--882.
\newblock
  \href{http://www.aclweb.org/anthology/P12-1092}{http://www.aclweb.org/anthology/P12-1092}.

\bibitem[{Jebara et~al.(2004)Jebara, Kondor, and Howard}]{prob_product_kernel}
Tony Jebara, Risi Kondor, and Andrew Howard. 2004.
\newblock Probability product kernels.
\newblock {\em Journal of Machine Learning Research\/} 5:819--844.

\bibitem[{Kim et~al.(2016)Kim, Jernite, Sontag, and Rush}]{char_lm}
Yoon Kim, Yacine Jernite, David Sontag, and Alexander~M. Rush. 2016.
\newblock Character-aware neural language models.
\newblock In {\em Proceedings of the Thirtieth {AAAI} Conference on Artificial
  Intelligence, February 12-17, 2016, Phoenix, Arizona, {USA.}\/}. pages
  2741--2749.

\bibitem[{Kuru et~al.(2016)Kuru, Can, and Yuret}]{char_ner}
Onur Kuru, Ozan~Arkan Can, and Deniz Yuret. 2016.
\newblock \href{http://aclweb.org/anthology/C/C16/C16-1087.pdf}{Charner:
  Character-level named entity recognition}.
\newblock In {\em {COLING} 2016, 26th International Conference on Computational
  Linguistics, Proceedings of the Conference: Technical Papers, December 11-16,
  2016, Osaka, Japan\/}. pages 911--921.
\newblock
  \href{http://aclweb.org/anthology/C/C16/C16-1087.pdf}{http://aclweb.org/anthology/C/C16/C16-1087.pdf}.

\bibitem[{Lee et~al.(2017)Lee, Cho, and Hofmann}]{char_mt_noseg}
Jason Lee, Kyunghyun Cho, and Thomas Hofmann. 2017.
\newblock
  \href{https://transacl.org/ojs/index.php/tacl/article/view/1051}{Fully
  character-level neural machine translation without explicit segmentation}.
\newblock {\em {TACL}\/} 5:365--378.
\newblock
  \href{https://transacl.org/ojs/index.php/tacl/article/view/1051}{https://transacl.org/ojs/index.php/tacl/article/view/1051}.

\bibitem[{Leviant and Reichart(2015)}]{it_de_rus_ws_sl}
Ira Leviant and Roi Reichart. 2015.
\newblock \href{http://arxiv.org/abs/1508.00106}{Judgment language matters:
  Multilingual vector space models for judgment language aware lexical
  semantics}.
\newblock {\em CoRR\/} abs/1508.00106.
\newblock
  \href{http://arxiv.org/abs/1508.00106}{http://arxiv.org/abs/1508.00106}.

\bibitem[{Luong et~al.(2013)Luong, Socher, and Manning}]{rarewords}
Minh-Thang Luong, Richard Socher, and Christopher~D. Manning. 2013.
\newblock Better word representations with recursive neural networks for
  morphology.
\newblock In {\em CoNLL\/}. Sofia, Bulgaria.

\bibitem[{Mikolov et~al.(2013{\natexlab{a}})Mikolov, Chen, Corrado, and
  Dean}]{word2vec}
Tomas Mikolov, Kai Chen, Greg Corrado, and Jeffrey Dean. 2013{\natexlab{a}}.
\newblock \href{http://arxiv.org/abs/1301.3781}{Efficient estimation of word
  representations in vector space}.
\newblock {\em CoRR\/} abs/1301.3781.
\newblock
  \href{http://arxiv.org/abs/1301.3781}{http://arxiv.org/abs/1301.3781}.

\bibitem[{Mikolov et~al.(2013{\natexlab{b}})Mikolov, Chen, Corrado, and
  Dean}]{word2vec2}
Tomas Mikolov, Kai Chen, Greg Corrado, and Jeffrey Dean. 2013{\natexlab{b}}.
\newblock \href{http://arxiv.org/abs/1301.3781}{Efficient estimation of word
  representations in vector space}.
\newblock {\em CoRR\/} abs/1301.3781.
\newblock
  \href{http://arxiv.org/abs/1301.3781}{http://arxiv.org/abs/1301.3781}.

\bibitem[{Mikolov et~al.(2011)Mikolov, Kombrink, Burget, Cernock{\'{y}}, and
  Khudanpur}]{rnnlm2}
Tomas Mikolov, Stefan Kombrink, Luk{\'{a}}s Burget, Jan Cernock{\'{y}}, and
  Sanjeev Khudanpur. 2011.
\newblock \href{https://doi.org/10.1109/ICASSP.2011.5947611}{Extensions of
  recurrent neural network language model}.
\newblock In {\em Proceedings of the {IEEE} International Conference on
  Acoustics, Speech, and Signal Processing, {ICASSP} 2011, May 22-27, 2011,
  Prague Congress Center, Prague, Czech Republic\/}. pages 5528--5531.
\newblock
  \href{https://doi.org/10.1109/ICASSP.2011.5947611}{https://doi.org/10.1109/ICASSP.2011.5947611}.

\bibitem[{Miller and Charles(1991)}]{evaldata_mc}
George~A. Miller and Walter~G. Charles. 1991.
\newblock \href{https://doi.org/10.1080/01690969108406936}{{Contextual
  Correlates of Semantic Similarity}}.
\newblock {\em Language \& Cognitive Processes\/} 6(1):1--28.
\newblock
  \href{https://doi.org/10.1080/01690969108406936}{https://doi.org/10.1080/01690969108406936}.

\bibitem[{Neelakantan et~al.(2014)Neelakantan, Shankar, Passos, and
  McCallum}]{nonparam_multiprototype}
Arvind Neelakantan, Jeevan Shankar, Alexandre Passos, and Andrew McCallum.
  2014.
\newblock \href{http://aclweb.org/anthology/D/D14/D14-1113.pdf}{Efficient
  non-parametric estimation of multiple embeddings per word in vector space}.
\newblock In {\em Proceedings of the 2014 Conference on Empirical Methods in
  Natural Language Processing, {EMNLP} 2014, October 25-29, 2014, Doha, Qatar,
  {A} meeting of SIGDAT, a Special Interest Group of the {ACL}\/}. pages
  1059--1069.
\newblock
  \href{http://aclweb.org/anthology/D/D14/D14-1113.pdf}{http://aclweb.org/anthology/D/D14/D14-1113.pdf}.

\bibitem[{Pennington et~al.(2014)Pennington, Socher, and Manning}]{glove}
Jeffrey Pennington, Richard Socher, and Christopher~D. Manning. 2014.
\newblock \href{http://aclweb.org/anthology/D/D14/D14-1162.pdf}{Glove: Global
  vectors for word representation}.
\newblock In {\em Proceedings of the 2014 Conference on Empirical Methods in
  Natural Language Processing, {EMNLP} 2014, October 25-29, 2014, Doha, Qatar,
  {A} meeting of SIGDAT, a Special Interest Group of the {ACL}\/}. pages
  1532--1543.
\newblock
  \href{http://aclweb.org/anthology/D/D14/D14-1162.pdf}{http://aclweb.org/anthology/D/D14/D14-1162.pdf}.

\bibitem[{Radinsky et~al.(2011)Radinsky, Agichtein, Gabrilovich, and
  Markovitch}]{evaldata_mturk287}
Kira Radinsky, Eugene Agichtein, Evgeniy Gabrilovich, and Shaul Markovitch.
  2011.
\newblock \href{http://doi.acm.org/10.1145/1963405.1963455}{A word at a time:
  Computing word relatedness using temporal semantic analysis}.
\newblock In {\em Proceedings of the 20th International Conference on World
  Wide Web\/}. WWW '11, pages 337--346.
\newblock
  \href{http://doi.acm.org/10.1145/1963405.1963455}{http://doi.acm.org/10.1145/1963405.1963455}.

\bibitem[{Rubenstein and Goodenough(1965)}]{evaldata_rg}
Herbert Rubenstein and John~B. Goodenough. 1965.
\newblock \href{http://doi.acm.org/10.1145/365628.365657}{Contextual correlates
  of synonymy}.
\newblock {\em Commun. ACM\/} 8(10):627--633.
\newblock
  \href{http://doi.acm.org/10.1145/365628.365657}{http://doi.acm.org/10.1145/365628.365657}.

\bibitem[{Smith et~al.(2017)Smith, Turban, Hamblin, and
  Hammerla}]{multilingual_emb}
Samuel~L. Smith, David H.~P. Turban, Steven Hamblin, and Nils~Y. Hammerla.
  2017.
\newblock \href{http://arxiv.org/abs/1702.03859}{Offline bilingual word
  vectors, orthogonal transformations and the inverted softmax}.
\newblock {\em CoRR\/} abs/1702.03859.
\newblock
  \href{http://arxiv.org/abs/1702.03859}{http://arxiv.org/abs/1702.03859}.

\bibitem[{Spearman(1904)}]{spearman04}
C.~Spearman. 1904.
\newblock The proof and measurement of association between two things.
\newblock {\em American Journal of Psychology\/} 15:88--103.

\bibitem[{Tian et~al.(2014)Tian, Dai, Bian, Gao, Zhang, Chen, and
  Liu}]{multi_word_embs}
Fei Tian, Hanjun Dai, Jiang Bian, Bin Gao, Rui Zhang, Enhong Chen, and
  Tie{-}Yan Liu. 2014.
\newblock \href{http://aclweb.org/anthology/C/C14/C14-1016.pdf}{A probabilistic
  model for learning multi-prototype word embeddings}.
\newblock In {\em {COLING} 2014, 25th International Conference on Computational
  Linguistics, Proceedings of the Conference: Technical Papers, August 23-29,
  2014, Dublin, Ireland\/}. pages 151--160.
\newblock
  \href{http://aclweb.org/anthology/C/C14/C14-1016.pdf}{http://aclweb.org/anthology/C/C14/C14-1016.pdf}.

\bibitem[{Vilnis and McCallum(2014)}]{word2gauss}
Luke Vilnis and Andrew McCallum. 2014.
\newblock \href{http://arxiv.org/abs/1412.6623}{Word representations via
  gaussian embedding}.
\newblock {\em CoRR\/} abs/1412.6623.
\newblock
  \href{http://arxiv.org/abs/1412.6623}{http://arxiv.org/abs/1412.6623}.

\bibitem[{Yang and Powers(2006)}]{evaldata_yp}
Dongqiang Yang and David M.~W. Powers. 2006.
\newblock Verb similarity on the taxonomy of wordnet.
\newblock In {\em In the 3rd International WordNet Conference (GWC-06), Jeju
  Island, Korea\/}.

\bibitem[{Zhao and Zhang(2016)}]{char_mt_efficient}
Shenjian Zhao and Zhihua Zhang. 2016.
\newblock \href{http://arxiv.org/abs/1608.04738}{An efficient character-level
  neural machine translation}.
\newblock {\em CoRR\/} abs/1608.04738.
\newblock
  \href{http://arxiv.org/abs/1608.04738}{http://arxiv.org/abs/1608.04738}.

\bibitem[{Zipf(1949)}]{zipf}
G.K. Zipf. 1949.
\newblock {\em Human behavior and the principle of least effort: an
  introduction to human ecology\/}.
\newblock Addison-Wesley Press.
\newblock
  \href{https://books.google.com/books?id=1tx9AAAAIAAJ}{https://books.google.com/books?id=1tx9AAAAIAAJ}.

\end{thebibliography}
\bibliographystyle{acl_natbib}
\end{document}